\documentclass[lettersize,journal]{IEEEtran}
\usepackage{amsmath,amsfonts}
\usepackage{algorithmic}
\usepackage{algorithm}
\usepackage{array}
\usepackage[caption=false,font=normalsize,labelfont=sf,textfont=sf]{subfig}
\usepackage{textcomp}
\usepackage{stfloats}
\usepackage{url}
\usepackage{verbatim}
\usepackage{graphicx}
\usepackage{booktabs}
\usepackage{cite}
\begin{document}


\title{Gene-MOE: A sparsely gated prognosis and classification framework exploiting pan-cancer genomic information}

\author{Xiangyu Meng, Xue Li, Qing Yang, Huanhuan Dai, Lian Qiao, Hongzhen Ding, Long Hao, Xun Wang
\thanks{Xiangyu Meng, Xue Li, Qing Yang, Huanhuan Dai, Lian Qiao, Hongzhen Ding, Long Hao and Xun Wang are with the Department of Computer Science and Technology, China University of Petroleum, Qingdao, Shandong, 266580, China (e-mail: x.meng@s.upc.edu.cn; Xueleecs@gmail.com; s21070069@s.upc.edu.cn; daihuanhuan0901@163.com; s21070055@s.upc.edu.cn; s22070050@s.upc.edu.cn; 1808010425@s.upc.edu.cn; wangsyun@upc.edu.cn) (\textit{Corresponding author: Xun Wang})}

}

\markboth{Journal of \LaTeX\ Class Files,~Vol.~14, No.~8, August~2021}%
{Shell \MakeLowercase{\textit{et al.}}: A Sample Article Using IEEEtran.cls for IEEE Journals}

\IEEEpubid{0000--0000/00\$00.00~\copyright~2021 IEEE}

\maketitle



\begin{abstract}
Deep learning based genomic analysis methods enhanced our understanding for cancer research. 
However, the overfitting issue, arising from the limited number of patient samples, presents a challenge in improving the accuracy of genome analysis by deepening the neural network.
Furthermore, it remains uncertain whether novel approaches such as the sparsely gated mixture of expert (MOE) and self-attention mechanisms can improve the accuracy of genomic analysis.
We introduce a novel sparsely gated prognosis and classification analysis framework called Gene-MOE.
This framework exploits the potential of the MOE layers and the proposed mixture of attention expert (MOAE) layers to enhance the analysis accuracy. Additionally, it addresses overfitting challenges by integrating pan-cancer information from 33 distinct cancer types through pre-training.
According to the survival analysis results on 14 cancer types, Gene-MOE outperformed state-of-the-art models on 12 cancer types. Through detailed feature analysis, we found that the Gene-MOE model could learn rich feature representations of high-dimensional genes.
According to the classification results, the total accuracy of the classification model for 33 cancer classifications reached 95.8\%, representing the best performance compared to state-of-the-art models.
These results indicate that Gene-MOE holds strong potential for use in cancer classification and survival analysis. 
\end{abstract}



\maketitle

\section{Introduction}
\label{sec:introduction}

Carcinogenic manifestations often result from mutations in one or more genes \cite{linares2021machine}. 
The Cancer Genome Atlas (TCGA) project represents a major advance in cancer genomics. 
The tens of thousands of pretreatment samples in TCGA, including over 30 cancer types and numerous measurements, including RNA sequencing (RNA-seq), DNA methylation, and copy-number variation, deepen our understanding of cancer-related genes and their clinical relevance \cite{zhang2019survey,li2023proteogenomic,icgc2020pan}. 
To find more meaningful biological functions, researchers have developed methods such as prognosis prediction \cite{zhu2020application, kourou2015machine}, tumor subtypes \cite{sorlie2003repeated}, microsatellite instability (MSI) \cite{hause2016classification}, immunological aspects \cite{sanchez2008cholera}, and certain pathways of interest \cite{wang2010analysing}.

In the past two decades, a series of machine learning methods have been proposed for genomic analysis. Representative work includes gene survival analysis using the COX regression model~\cite{lin1989robust}, cancer classification and tumor biomarker identification using random forest \cite{capper2018dna,diaz2006gene}, cluster gene expression data using principal component analysis (PCA) \cite{yeung2001principal}, gene selection and cancer classification using support vector machine (SVM) \cite{duan2005multiple}, and gene expression pattern classification using the linear regression method \cite{liu2019prediction}.
Owing to their simplicity and user-friendliness, machine learning models have proved to be highly efficient in processing genomic analysis tasks, delivering commendable results across various functions, such as dimensional reduction and clustering visualization. However, the nonlinear biological characteristics inherent in high-dimensional genes cannot be effectively learned by the machine learning method, leading to precision loss in genomic analysis.

In recent years, deep learning methods have received widespread attention. With powerful fitting capabilities and more accurate prediction effects than traditional learning methods, deep learning methods can more effectively handle the potential correlation features of high-dimensional data. Many researchers have adopted deep learning methods for feature extraction and downstream prediction. Representative work includes the Cox-nnet model \cite{ching2018cox}, DeepCC~\cite{gao2019deepcc}, DeepCues \cite{zeng2021deep}, PathCNN \cite{oh2021pathcnn}, and the cancer prognosis and classification method using a graph convolutional network (GCN) \cite{ramirez2020classification, ramirez2021prediction}.
Unlike machine learning methods, deep learning methods have greatly improved fitting performance and prediction accuracy, thus improving cancer diagnosis and prognosis. However, owing to the imbalance of samples between different cancer types and the few fitting samples for cancer types such as uterine carcinosarcoma (UCS), ocular melanoma (UVM), and large b-cell lymphoma (DLBC), deep learning methods must build two to four layers of a deep neural network (DNN) to prevent over-fitting problems, so the model cannot learn the deep correlations between high-dimensional genes.
\IEEEpubidadjcol
Recently, the pre-trained models have received widespread attention owing to their powerful feature learning capabilities. These models require training a network with billions of parameters to learn common features across various data samples and transferring weights to a specific task to capture the unique properties. Some notable models are transformer-based models \cite{li2023marppi,wang2022transfusionnet}, which were first proposed in the natural language processing field. They include dense models such as GPT-3 \cite{brown2020language}, Gopher \cite{rae2021scaling}, and the sparse model based on the mixture of expert (MOE) model \cite{DBLP:conf/iclr/ShazeerMMDLHD17, DBLP:conf/iclr/LepikhinLXCFHKS21}. Moreover, many pre-training models have been introduced to solve biological problems and have achieved remarkable results, including BioBERT \cite{lee2020biobert}, DNABERT \cite{ji2021dnabert}, and scBERT \cite{yang2022scbert}. 

However, challenges persist in the application of pre-trained models to high-dimensional genetic data.
First, the number of genes far exceeds the number of patient samples, and most genes contain no useful information for diagnosis and prognosis, exacerbating the risk of over-fitting during training. The TCGA dataset is estimated to contain over 50,000 genes, of which protein-coding genes make up approximately 21,000 \cite{pertea2018thousands}, and most genes contain no useful information.
Appropriate data preprocessing and augmentation methods must be designed to avoid meaningless information as much as possible.
Second, a variety of genes often express together to cause cancer, and a strong correlation exists between different genes. Exploring such correlations from high-dimensional genetic features requires special consideration to design appropriate feature extraction methods tailored to genomic data. 

In this work, we combined the principles of the MOE structure to create a pre-trained feature extraction model called Gene-MOE for high-dimensional RNA-seq gene expression data. This model combines the characteristics of MOE and uses 500 million parameters to fully learn the deep correlation features of high-dimensional genes and fit low-dimensional spatial features through the proposed unsupervised training strategy. We selected two common applications to use Gene-MOE for downstream prediction. According to the result, the Gene-MOE model achieved good performance for both types of tasks. Specifically, the main contributions were as follows:
\begin{enumerate}
\item We propose a sparsely gated RNA-seq analysis framework called Gene-MOE. Gene-MOE exploits the MOE layers to extract the features from high-dimensional RNA-seq genes. Furthermore, the self-attention mechanism is added to construct the MOAE model to further learn the deep semantic relationship inside the genetic features.
Finally, we use a novel self-supervised pre-training strategy to make Gene-MOE learn the common features of 33 cancers and then transfer the pre-trained weight to the specific analysis including survival analysis and cancer classification;
\item According to the survival analysis results on 14 cancer types, the Gene-MOE achieved the best concordance index on 12 cancer types. Moreover, the classifier using pre-trained Gene-MOE achieved accurate classification of 33 cancer types, with a total accuracy of 95.8\%;
\item According to the correlation analysis result, we found many cancer-related genes were strongly correlated with hidden nodes with high variance, which proves that Gene-MOE can effectively learn the gene expression of 33 types of cancer during the pre-training stage;
\item According to the visualization analysis, we found that the reconstructed genes coincided with the input genes during the pre-training phase, proving that Gene-MOE can perfectly learn the feature representation of high-dimensional genes.

\end{enumerate}

The rest of this paper is organized as follows. Section \ref{method} illustrates our method, including the dataset preparation, Gene-MOE framework illustration, and training strategy. Section \ref{result} presents our experimental results for Gene-MOE. Finally, in Section \ref{conclusion}, we provide our conclusion.

\section{Methods}

 \label{method}
\label{sec:headings}
\subsection{Dataset Preparation}
\begin{figure}[!t]
    \centering
    \centerline{\includegraphics[width=0.95\columnwidth]{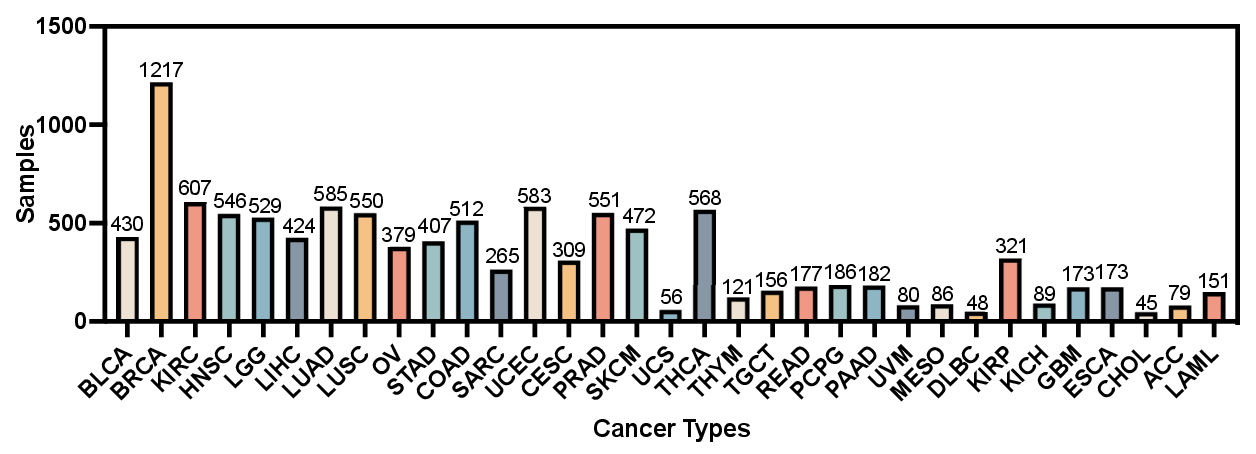}}
    \caption{Patient statistics of 33 cancer types after preprocessing.}
    \label{fig:datashow}
\end{figure}
In this study, we primarily utilized the pan-cancer RNA-seq database from the TCGA dataset of the Pan-Cancer Atlas project, which consists of 33 cancer types. 
Furthermore, we downloaded the specific RNA-seq dataset for each TCGA cancer type.
UCSC Xena already includes the preliminary processing for these 34 datasets, so we directly used the RNA-seq data after the FPKM-UQ normalization from UCSC Xena.
In the initial dataset, each patient was associated with 60,484 genes, a majority of which exhibited null expression and had no relevance to cancer. Consequently, our first step involved filtering out these non-contributing genes.
We first deleted the empty genes and selected overlapping genes among 34 datasets. 
Next, we filtered out genes with variances less than 0.4 and mean values less than 0.8 according to \cite{ramirez2021prediction}. The reason is that genes that meet this low variance and low mean range were 0 in most patients and did not make any contribution to model training.
Finally, before feeding the data into the network, we performed min-max normalization.
After preparation, we selected 25,182 genes for each dataset.
Figure \ref{fig:datashow} shows the patient number of these 33 cancer types.


\subsection{Framework Illustration}

\begin{figure*}[!t]
    \centering
    \centerline{\includegraphics[width=0.95\textwidth]{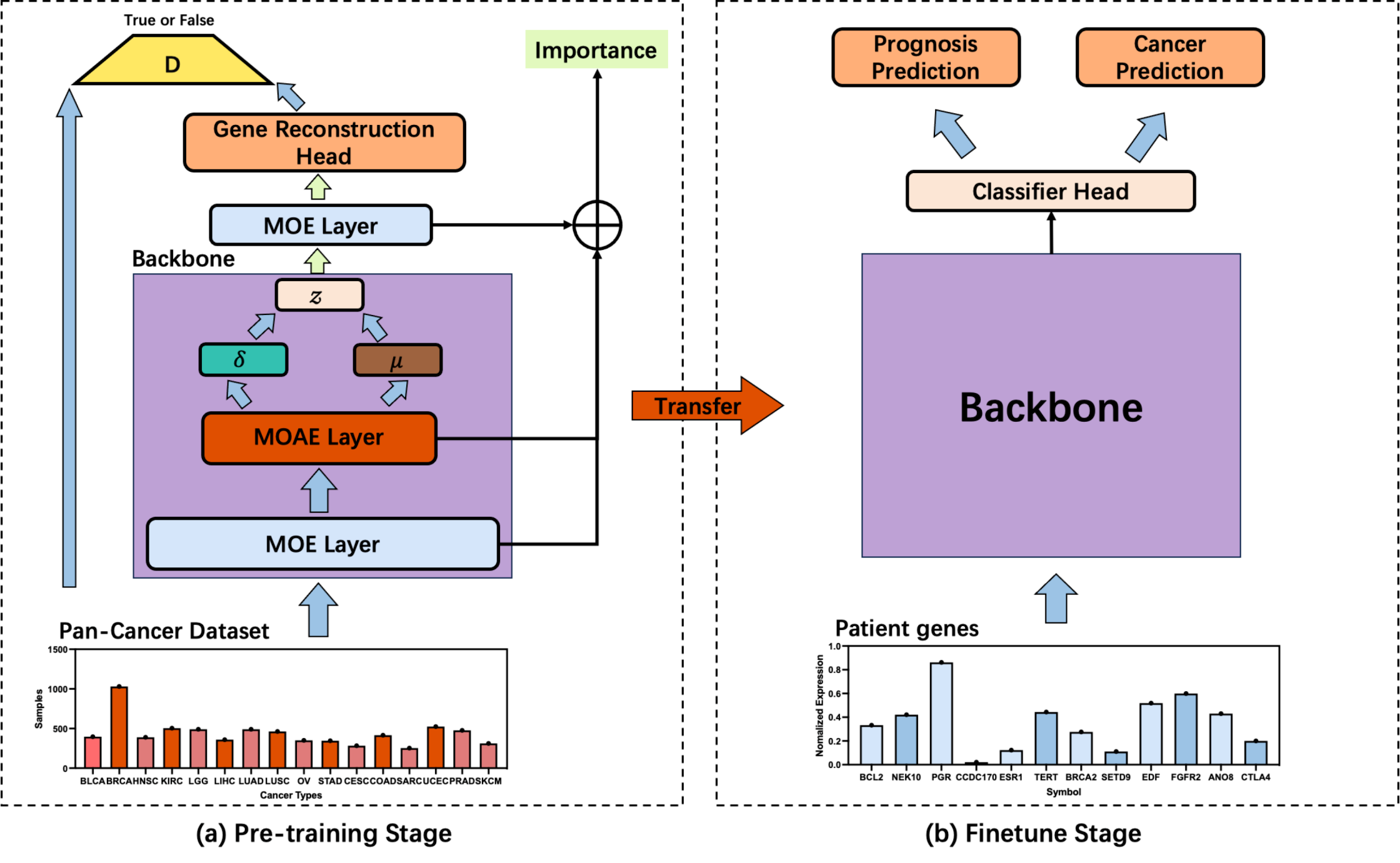}}
    \caption{Framework illustration of Gene-MOE. This model primarily consists of an encoder backbone network and a decoder network. (a) Pre-training stage. During this phase, Gene-MOE takes preprocessed pan-cancer genes as input to train the encoder--decoder network. (b) Fine-tuning stage. During this stage, Gene-MOE transfers the pre-trained backbone and connects to a classification head to achieve accurate downstream prediction tasks such as prognosis prediction, cancer classification, and specific pathways of interest.}
    \label{fig:model_frame}
\end{figure*}

Figure \ref{fig:model_frame} illustrates the framework of Gene-MOE, which comprises two stages: pre-training and fine-tuning. 
During the first stage, we construct the Gene-MOE model to learn low-dimensional feature encoding of high-dimensional pan-cancer genes. In this stage, we employ a self-supervised learning approach where the training labels are the input genes, aiming to acquire low-dimensional feature encoding of high-dimensional pan-cancer genes. The Gene-MOE model primarily consists of an encoder backbone network and a decoder network. Within the encoder network, we introduce an innovative MOE model based on sparse gating. This method involves multiple experts, enabling the encoder to learn rich feature representations of the genomic information. Moreover, we designed a MOAE model, which employs multiple attention mechanisms as distinct experts and uses a learnable sparse gating mechanism to merge attention features adaptively.
In the second stage, the backbone of the Gene-MOE is transferred, and a new classification head is constructed to accomplish a fast, accurate downstream task after fine-tuning.
This methodology enables us to fully leverage the Gene-MOE model in two stages, facilitating the learning of rich pan-cancer gene feature encoding and the achievement of excellent performance in multiple downstream tasks.

\subsubsection{Sparsely Gated MOE Layer} 
Each patient has over 20,000 genes, among which intricate correlations exist. 
In previous work, it was common to build one or multiple fully connected networks (FCNs) to learn and extract key features related to genes.
However, directly employing FCNs cannot effectively help the model learn these intricate correlations of high-dimensional gene input. The learned features would result in a substantial loss. In contrast to dense layers, the MOE module trains $N$ experts, each of which independently learns and extracts features based on the characteristics of the input data. 
Compared with FCN, the MOE model can integrate diverse features from multiple expert models, enhancing the feature extraction capabilities of high-dimension genes and the overall model performance. Moreover, the training of each expert is independent, allowing for adjustments based on the characteristics of gene data, thus providing greater flexibility compared with fully connected layers.
The MOE layer with $N$ experts is denoted as 
\begin{equation}
y = \sum_{i=1}^n\cdot G_i(x) \cdot D_i(x),
\end{equation}
where $D_i(x)$ is the expert network, which is an independent dense layer, and $G_i(x)$ is the sparsity gating network. Throughout the training phase of the model, $G_i(x)$ is employed to dynamically select the top $K$ experts, expressed as 
\begin{equation}
G(x) = Softmax(TopK(H(x),k)),
\end{equation}
where $TopK(H(x),k)$ denotes a discrete function that maps the input feature $H(x)$ to a mask $m \in \mathbb{R}^n$. It is denoted as
\begin{equation}
\label{eq:topk}
\text{TopK}(H(x), k)_i = \left\{
\begin{aligned}
    & H_i(x) && \text{if } H_i(x) \text{ in the top } K \text{of } v.\\
    & - \infty && \text{otherwise.}
\end{aligned}
\right.
\end{equation}
This equation generates $m$, in which the values in $H(x)$ corresponding to elements outside the top $K$ are set to $- \infty$ so that when feeding $m$ to softmax function, the corresponding elements are set to 0. Furthermore, to improve the load balance, we introduce the noise term to the input $x$ and generate $H(x)$, denoted as 
\begin{equation}
    H_i(x) = (x\cdot W)_i + z\cdot Softplus((x \cdot W_{noise})_i),
\end{equation}
where $z \sim N(0,1)$ is a random Gaussian noise, $W$ is a trainable weight matrix that learns the sparsity gating feature, and $W_{noise}$ is the weight that controls the noise increment.

\begin{figure}[!t]
    \centering
    \centerline{\includegraphics[width=0.95\columnwidth]{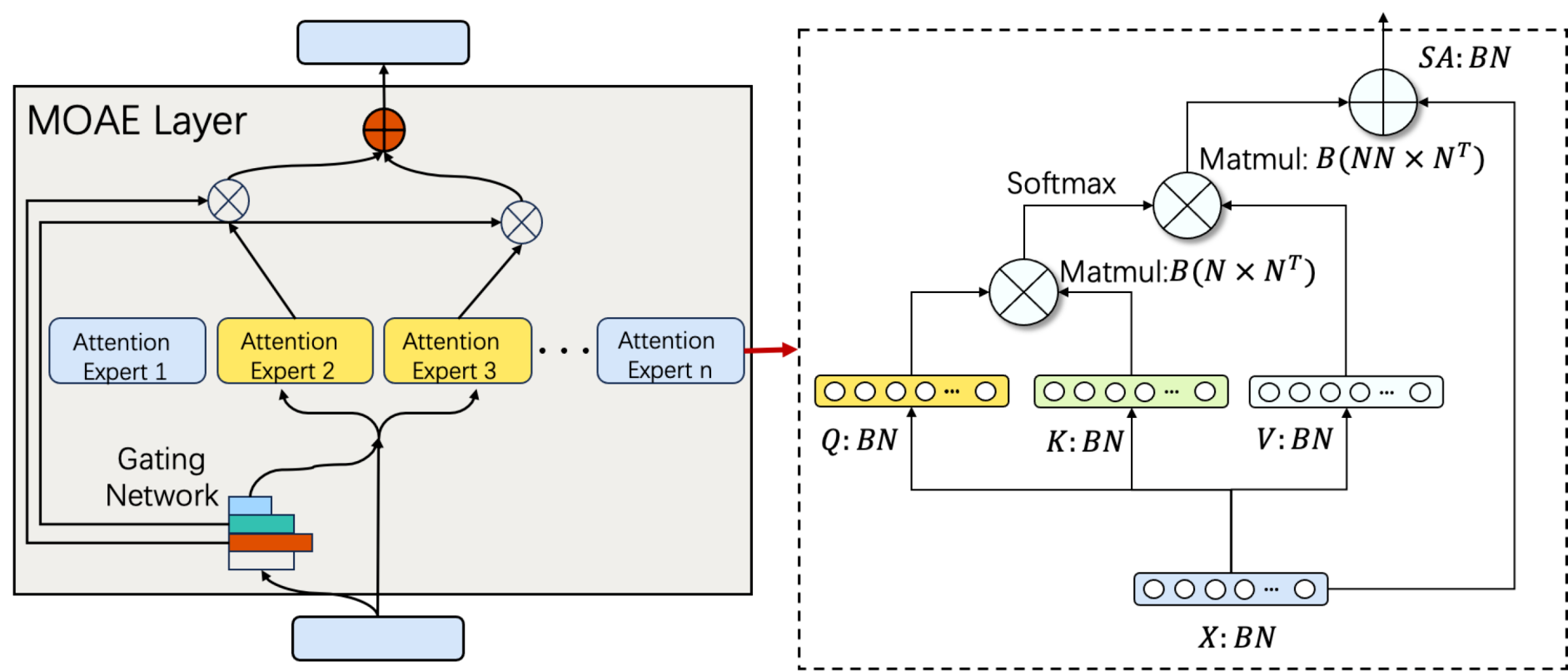}}
    \caption{Framework of Mixture of Attention Expert Layer (MOAE) layer}
    \label{fig:moae}
\end{figure}


\subsubsection{Mixture of Attention Expert Layer}
Using self-attention mechanisms can effectively help the model learn deep semantics in high-dimensional features. However, owing to the excessively high dimensions of genetic features, the direct use of self-attention methods still leads to some loss of crucial information. Moreover, given the scarcity of samples, stacking multiple layers of self-attention modules may introduce over-fitting issues. 
Therefore, we propose a mixture of attention expert (MOAE) model based on the MOE mechanism by constructing various attention experts and letting the model automatically choose the top $K$ attention experts with better effects. Training this module effectively fuses features extracted by the top-k self-attention experts, yielding richer semantic information. Unlike MOE, each expert in this module is a residual self-attention network, which is shown in Figure \ref{fig:moae}.

\subsection{Training Strategy}

We constructed a self-supervised pre-training strategy to learn the common features of pan-cancer genes and improve the feature extraction performance of the backbone network. 

\subsubsection{Data Augmentation by Gaussian Noise}
The Pan-cancer dataset had 10,000 patients after preprocessing. If we were to use 0.5 billion parameters, Gene-MOE might cause over-fitting issues. Therefore, we built a data augmentation strategy by introducing Gaussian noise. For an input gene $x$, the augmented input $\hat{x}$ is expressed as 
\begin{equation}
    \hat{x} = x + z,
\end{equation}
where $z \sim N(0,0.2)$. Moreover, the dropout strategy was introduced during the training.

\subsubsection{Joint Training Using a Generative Adversarial Network}

We used the training strategy of a generative adversarial network (GAN) to pre-train Gene-MOE. We constructed a discriminator called $D$, which is a dense layer network, and the Gene-MOE model is the generator. Then, Wasserstein loss was introduced to train $G$ and $D$ jointly. For $G$ and $D$, the loss is denoted as 
\begin{equation}
\begin{aligned}
    \mathcal{L}_{gan} = \mathbb{E}_{\hat{x} \sim P_{data}(\hat{x})} D(\hat{x}) - \mathbb{E}_{\hat{x} \sim P_{data}(\hat{x})} D(G(\hat{x})) -\\ \lambda_{gp} \mathbb{E}_{\hat{x} \sim \mathcal{X}}||||\nabla_{\hat{x}}D(\hat{x})||_2-1||_2,
    \label{loss_wgan}
\end{aligned}
\end{equation}
where $\lambda_{gp}$ is the hyper-parameter of the gradient penalty, and $\mathcal{X}$ is the sample space of $x$ and $\hat{x}$. Training using Equation \ref{loss_wgan} can help Gene-MOE learn how to perform dimensional reduction and reconstruct the generated gene.

\subsubsection{Measuring the Distribution } 
We further introduced the KL divergence to measure the similarity of the latent code $z$ generated by $\hat{x}$ and the standard Gaussian distribution. It can be denoted as 
\begin{equation}
    \mathcal{L}_{KL} = \sum_{i=0}^{n}(\mu(\hat{x})^2+\sigma(\hat{x})^2-log(\sigma(\hat{x})^2) -1),
    \label{KL}
\end{equation}
where $n$ represents the dim of $z$. Using this loss allows $z$ to maintain a standard normal distribution, thus simplifying the training difficulty of Wasserstein loss.

\subsubsection{Measuring the Similarity of Genes }
We introduced L1 loss to further measure the similarity of each gene between samples reconstructed by Gene-MOE and the ground-truth samples. It is computed as 
\begin{equation}
    \mathcal{L}_{L1} = ||G(\hat{x})-\hat{x}||_1.
    \label{L1}
\end{equation}

\subsubsection{Balancing Expert Utilization}
To allow each MOE layer to select each expert in a balanced manner, we introduced importance loss to each sparse gating layer of the MOE. The importance loss can be denoted as 
\begin{equation}
    \mathcal{L}_{importance} =  ||Importance(f(\hat{x}))||_2,
    \label{importance}
\end{equation}
where $f(\hat{x})$ denotes the input features of the gating layer, and $Importance(f(\hat{x}))$ denotes 
\begin{equation}
    Importance(f(\hat{x})) = \sum_{i \in B} G(f(\hat{x})_i),
    \label{wimportance}
\end{equation}
where $B$ denotes the batch size of the input features, and $G$ denotes the gating layer. To further improve the balanced loading, we also introduced the load balance loss $\mathcal{L}_{load}$ in \cite{DBLP:conf/iclr/ShazeerMMDLHD17}.

\subsubsection{Overall Pre-training Loss}
Combining these losses, we can express the overall loss of MOE as 
\begin{equation}
\begin{aligned}
    \mathcal{L}_{total} = \mathcal{L}_{gan} + \lambda_{KL} \cdot \mathcal{L}_{KL} + \lambda_{l1} \cdot \mathcal{L}_{L1} +\\ \lambda_{balance}\cdot (\mathcal{L}_{importance} + \mathcal{L}_{load}),
    \label{total}
\end{aligned}
\end{equation}
where $\lambda_{KL}$, $\mathcal{L}_{KL}$, $\lambda_{l1}$, and $\lambda_{balance}$ are the hyper-parameters. Therefore, the pre-training stage aims to fit the optimal parameters $\theta^{\ast}_G$ of MOE by solving 
\begin{equation}
    \theta^{\ast}_G = arg \mathop{min}\limits_{G} \mathop{max}\limits_{D} \mathcal{L}_{total}.
    \label{solve}
\end{equation}

\subsection{Experimental Settings}
The Gene-MOE was implemented using the PyTorch framework. We trained and evaluated Gene-MOE using an NVIDIA Tesla V100 (32GB) GPU. 
During the pre-training stage, we initially trained the Gene-MOE model on the normalized pan-cancer dataset including 33 cancer types. We then randomly divided the pan-cancer data into train dataset and test dataset as a ratio of 4:1, where the train dataset was used for training and the test dataset was used for feature analysis.
The Adam optimizer was used in the Gene-MOE training. During the pre-training stage, the learning rate of the optimizer was 0.0002, the total epochs were 200, and the batch size was 256. The hyper-parameters of the loss function were $\lambda_{kl} = 10, \lambda_{l1} = 20, \lambda_{balance}=10, \lambda_{gp} = 10$. For better convergence, we introduced a learning rate decay method that sets the learning rate that remains constant for the first 100 epochs and decays linearly to 0 for the last 100 epochs.

After the pre-training phase, we combined the backbone of the Gene-MOE with the classifier to perform survival analysis and cancer classification tasks.
During the survival analysis phase, we evaluated the survival model in the same way as Cox-nnet \cite{ching2018cox} and the VAE-Cox \cite{kim2020improved}. We trained specific Gene-MOE on 14 TCGA datasets of common cancer types. For each dataset, we used the same way to randomly divide the train dataset and test dataset as a ratio of 4:1. The optimal hyperparameters of each model were selected using the Bayesian Optimization strategy. Moreover, we repeated the entire process 5 times and calculated the average result to avoid the bias of the random splitting. In the cancer classification phase, due to the imbalance of samples for each cancer, directly dividing the pan-cancer data into the training set and the test set at a ratio of 4:1 will result in the missing cancer types with a small number of samples in the test set. To solve the above issues, we introduced a new partition method that the sample set of each cancer was divided into the train subset and test subset according to the ratio of 4:1, and then the training subsets and test subsets of 33 cancers were combined to obtain the training set and test set of the classification task.
We also used the Bayesian Optimization strategy to search the best hyperparameters and repeated the entire classification process 5 times to calculate the average result.

\begin{figure}[!t]
    \centering
    \centerline{\includegraphics[width=0.95\columnwidth]{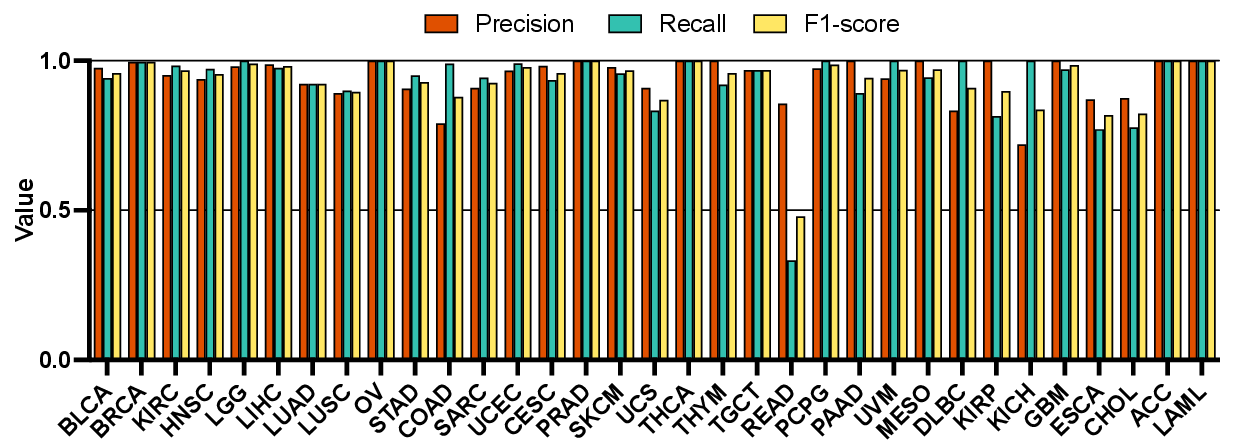}}
    \caption{Classification performance on 33 cancer types using Gene-MOE.}
    \label{fig:class}
\end{figure}
\subsection{Evaluation Metric}

In the survival analysis phase, the evaluation method we mainly used was the Concordance Index \cite{steck2007ranking}, which is widely used in survival analysis models and ranges from 0 to 1. When the Concordance Index $\leq$ 0.5, the model has completed an ineffective survival analysis prediction. When the Concordance index $\geq$ 0.5 and higher, the prediction effect of the model has been better.
In the cancer classification phase, we mainly used the accuracy, precision, recall, and F1-score. 
The accuracy is denoted as 
\begin{equation}
Acc = \frac{1}{c}\sum_i^{c} \frac{TP_i+TN_i}{N_i},
\end{equation}
where $c$ denotes the number of the class, $TP_i$ denotes the true positive samples of class $i$, $TN_i$ denotes the true negative of class $i$, and $N_i$ represents the total samples of class $i$. Precision metrics express the ability of the classifier to correctly predict the accuracy of positive samples, and it is denoted as
\begin{equation}
Precision = \frac{1}{c} \sum_i^{c} \frac{TP_i}{TP_i+FP_i},
\end{equation}
where $FP_i$ denotes the false positive samples of class $i$. The recall metric reflects the ability of the classifier to correctly predict the fullness of positive samples, and it is denoted as
\begin{equation}
Recall = \frac{1}{c} \sum_i^{c} \frac{TP_i}{TP_i+FN_i},
\end{equation}
where $FN_i$ denotes the false negative samples of class $i$. Finally, the F1-score denotes the harmonic mean of precision and recall, and it is expressed as  
\begin{equation}
F1_{score} = 2 \times \frac{Precision \times Recall}{Precision + Recall}.
\end{equation}

\section{Results} \label{result}



\subsection{Performance on Survival Analysis}

We chose COX survival analysis as a case to evaluate the effectiveness of Gene-MOE. We employed the backbone of Gene-MOE and loaded pre-trained weights. Subsequently, we integrated Gene-MOE with the COX classification head to train the survival analysis model. The training strategy of survival analysis adopted the Cox-ph model, which is denoted as
\begin{equation}
    h(t|x_i) = h_0(t)exp(\theta^T \cdot x_i),
    \label{cox}
\end{equation}
where $h_0(t)$ is the baseline hazard function, $\theta^T$ refers to the trainable parameters of the Cox model, and $x_i$ represents the hazard ratio of patients, which is the low-dimensional feature generated by the backbone of Gene-MOE. Training the Cox model was aimed at solving
\begin{equation}
    \theta^{\ast} = arg \mathop{min}\limits_{\theta} \sum_{C(i)=1}(\theta^T \cdot x_i - log \sum_{t_j \geq t_i} \theta^T \cdot x_j ),
    \label{cox2}
\end{equation}
where $t$ is the survival time of the patient sample, and $C(i)$ indicates whether the patient sample $i$ is censored. 
For each cancer type, we trained the individual survival model and evaluated it to prove its accuracy.

\begin{figure*}[!t]
    \centering
    \centerline{\includegraphics[width=0.95\textwidth]{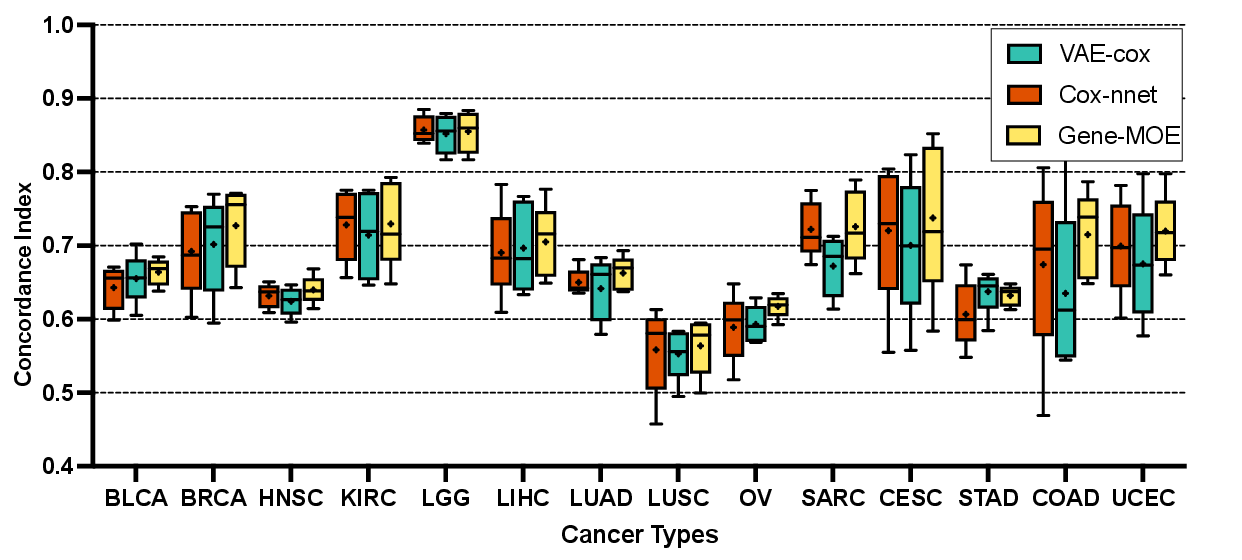}}
    \caption{Concordance Index Comparison of survival analysis on 14 cancer types. "$+$" of each box represents the mean Concordance Index.}
    \label{fig:cindex}
\end{figure*}

\begin{figure*}[!t]
    \centering
    \centerline{\includegraphics[width=0.95\textwidth]{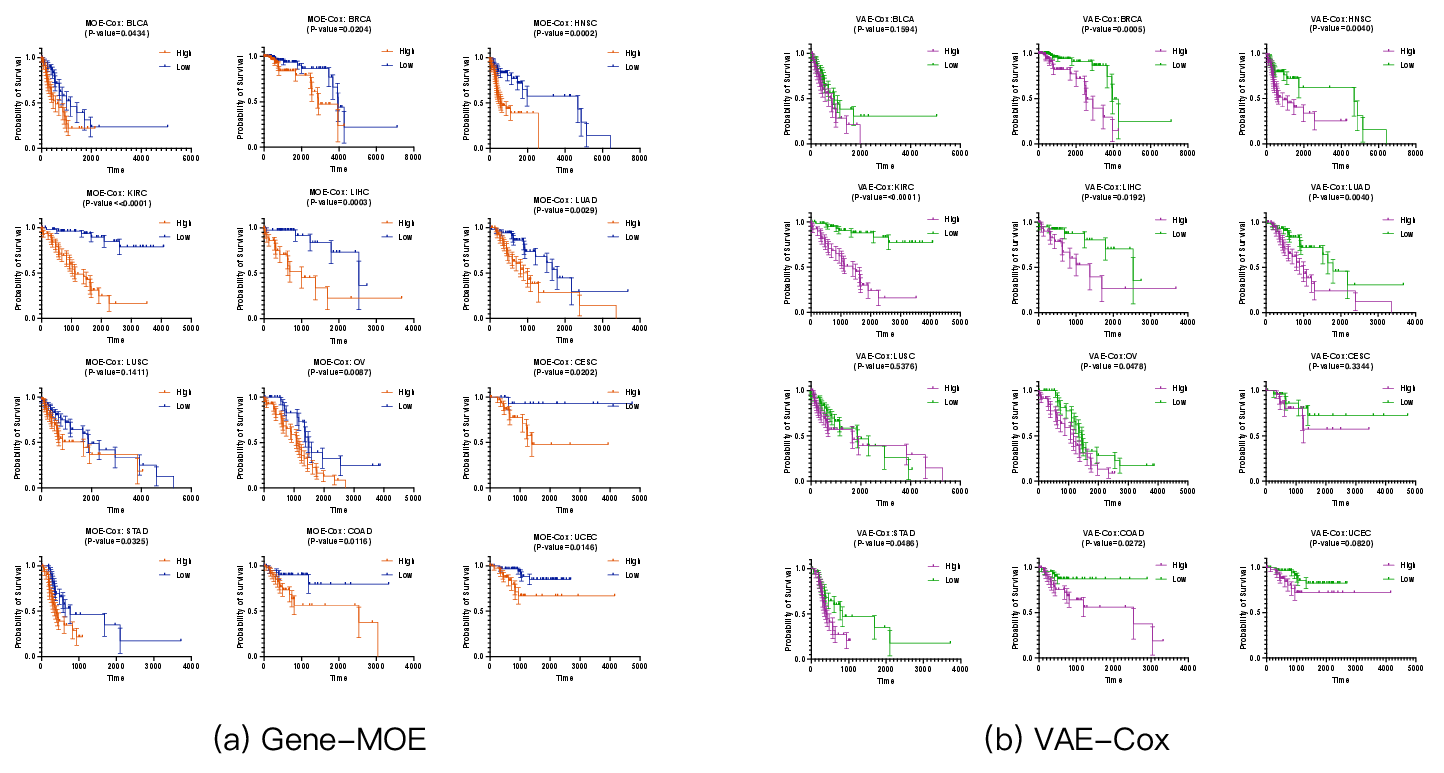}}
    \caption{Kaplan--Meier survival curves using MOE-Cox and VAE-cox on 12 cancer types. The Gene-MOE model shows a lower logP value on 10 cancer data sets. It shows that the Gene-MOE can predict hazard ratios that divide into high-risk groups and low-risk groups more significantly compared with VAE-cox on 10 cancer datasets.}
    \label{fig:km}
\end{figure*}
We selected 14 representative cancer types and analyzed the performance of our method in comparison with two state-of-the-art models: Cox-nnet \cite{ching2018cox} and VAE-Cox \cite{kim2020improved}.
Figure \ref{fig:cindex} shows the Concordance Index on 14 cancer types. Compared with Cox-nnet and the VAECox, the Gene-MOE outperformed on 12 cancer types, with a higher mean Concordance Index, which shows that Gene-MOE carried out a more accurate survival analysis compared with these two models.

We further employed Gene-MOE for survival analysis predictions. Specifically, we selected the test datasets of the 12 cancer types indicated in Figure \ref{fig:cindex}. Patients were divided into high-risk and low-risk groups based on the average predictions of Gene-MOE. Subsequently, we plotted Kaplan--Meier (KM) survival curves and conducted log-rank tests. We also carried out the same survival analysis for VAE-Cox. Figure \ref{fig:km} displays the KM curve results for 12 cancer types.
We found that Gene-MOE outperformed VAE-Cox significantly for these 12 cancer types with a lower logP value, indicating the effective capability of Gene-MOE to split patients into high-risk and low-risk groups.

\subsection{Performance on Cancer Type Classification}

We assessed the performance of Gene-MOE through a cancer classification task. 
Specifically, we designed a new classification head that takes the Gene-MOE's backbone network as input, predicting probabilities for 33 distinct cancer types. 
Then Focal loss~\cite{lin2017focal} was employed during the training process to mitigate the imbalance issue and enhance the model performance. 
Subsequently, we evaluated the precision, recall, and F1-score of the classification model employing Gene-MOE. The performance results are depicted in Figure \ref{fig:class}. As illustrated in this figure, our model demonstrated significant performance on 32 cancer types except rectum adenocarcinoma (READ), with a total accuracy of 95.8\%. 

To prove the novelty of proposed classifier, we selected several state-of-the-art classifiers for comparison, including classifier models based on machine learning methods such as random forest \cite{statnikov2008comprehensive} and SVM \cite{kourou2015machine,statnikov2008comprehensive}, as well as deep learning based classifier models like MLP \cite{ahn2018deep} and CNN \cite{mostavi2020convolutional}. 
To ensure the fairness of the comparative experiments, we conducted five independent processes and calculated the average of the classification results to avoid the bias. Table \ref{tab:class_compare} shows the classification result of these five models. According to table \ref{tab:class_compare}, We found that the classification results of the Gene-MOE model on the test dataset was improved by 0.1-0.5 compared with the other four methods.

\begin{table}[t]
\caption{Comparison of classification metrics}%
\begin{tabular*}{0.95\columnwidth}{@{\extracolsep\fill}lcccc@{\extracolsep\fill}}
\toprule
  & Accuracy & Precision & Recall & F1-Score \\ \midrule
RandomForest         & 0.9010 &   0.8943 & 0.8338       &  0.8409        \\
SVM         &  0.9498        &  0.9281         & 0.9114       & 0.9149         \\
MLP       &  0.9250        &  0.8989     &  0.8756      &   0.8790       \\
2D-Hybrid-CNN & 0.9507    &  0.9300     & 0.9229  & 0.9232         \\ 
Gene-MOE & \textbf{0.9580}    &  \textbf{0.9554}     & \textbf{0.9374}  & \textbf{0.9333 }        \\ 
\bottomrule
\end{tabular*}
\label{tab:class_compare}
\end{table}

\begin{figure}[!t]
    \centering
    \centerline{\includegraphics[width=\columnwidth]{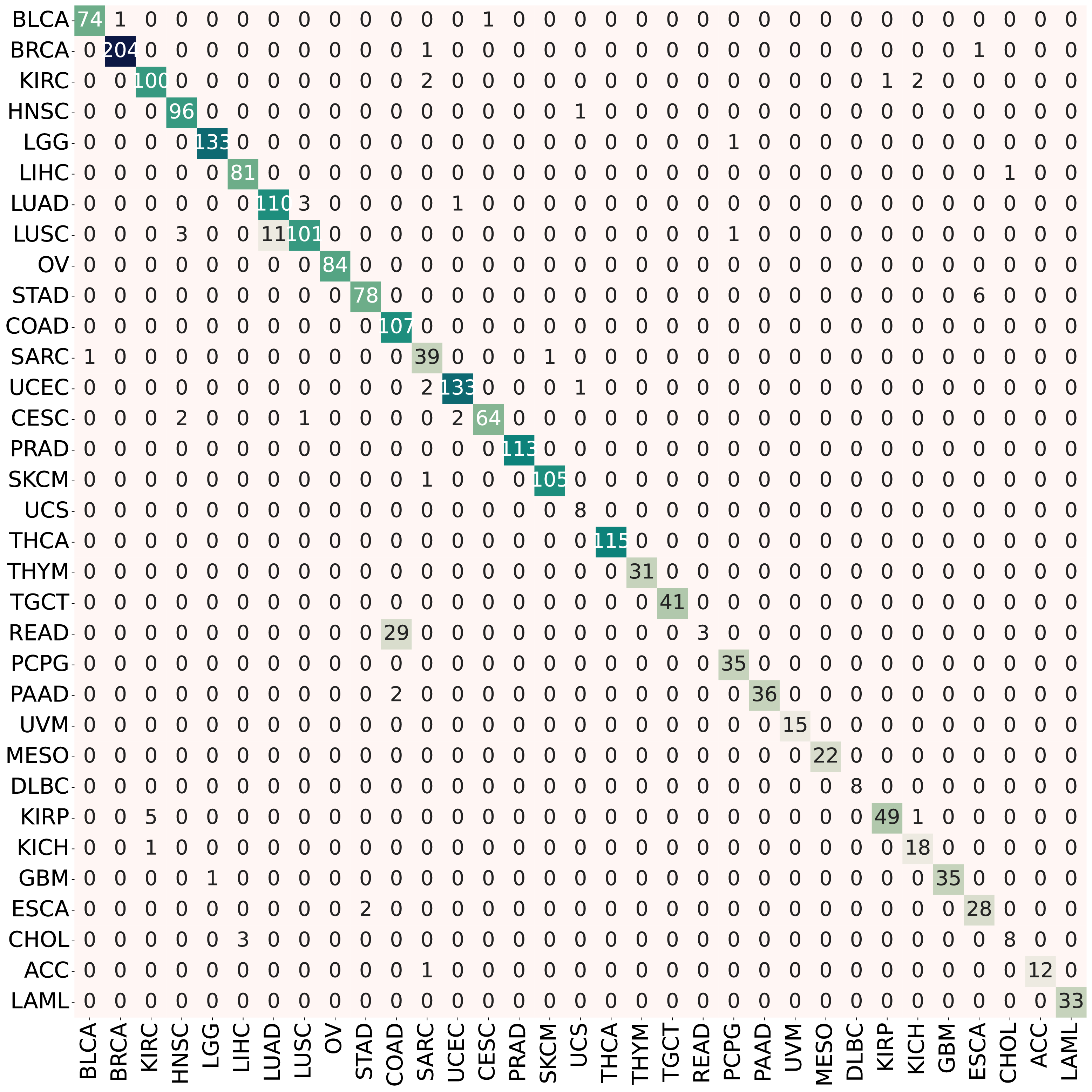}}
    \caption{Confusion matrix of test samples predicted by classification model using Gene-MOE with 33 cancer types.}
    \label{fig:confusion}
\end{figure}

We further analyzed the classification performance by constructing a confusion matrix. Figure \ref{fig:confusion} illustrates the confusion matrix constructed for 33 types of cancer. Our model demonstrated accurate classification of all 33 cancer types with a notably low misclassification rate according to the confusion matrix. Moreover, we observed that our model misclassified 24 cases of READ as colon adenocarcinoma (COAD). This issue occurred because READ and COAD are genetically identical \cite{cancer2012comprehensive}, and because COAD samples outnumbered READ samples, causing the model to misclassify some READ patients as COAD patients.

\begin{figure*}[!t]
    \centering
    \centerline{\includegraphics[width=0.9\textwidth]{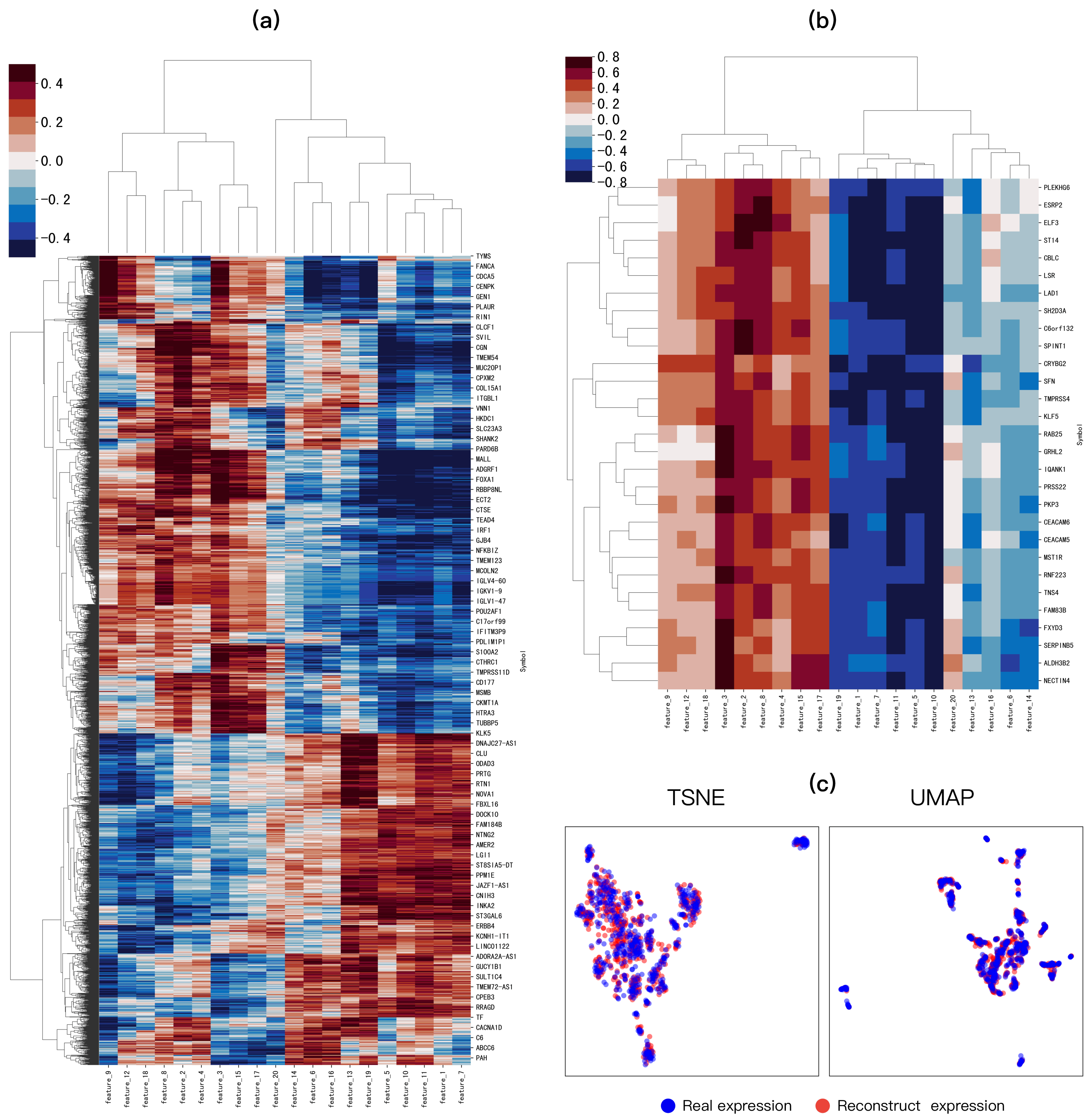}}
    \caption{Feature analysis of Gene-MOE. (a) Pearson correlation heat map between the leading feature and the patient genes. The features learned by Gene-MOE were strongly correlated with patient genes. (b) Pearson correlation heat map by selecting mean absolute coefficient greater than $0.5$. (c) TSNE and UMAP results of real genes and Gene-MOE reconstructed genes. According to the results, the reconstructed genes maintained the same distribution as the real genes.}
    \label{fig:feature}
\end{figure*}

\subsection{Feature Analysis of Gene-MOE}
\subsubsection{Correlation Analysis}
We conducted a correlation analysis between the low-dimensional features extracted by the Gene-MOE model and the high-dimensional genes of patients. Specifically, we used 1,000 patient samples from the test dataset to feed into the Gene-MOE model, and the model evaluated the low-dimensional features of each patient. We then selected the top 20 features with the highest variance as the leading features and calculated their Pearson correlations with the original patient genetic information.
The results are shown in Figure~\ref{fig:feature}(a), which presents a heat map illustrating the correlation between these leading features and the genes of 1,000 patients, which unequivocally indicates a significant correlation between the low-dimensional features predicted by the Gene-MOE model and the original input features.
Based on Figure~\ref{fig:feature}(a), we further refined our analysis by identifying genes with an average absolute correlation exceeding 0.4 with respect to the 20 leading features. 
Then, 29 strongly correlated genes were filtered, which is shown in Figure~\ref{fig:feature}(b). According to the result in Figure~\ref{fig:feature}(b), we observed that many genes with strong correlations to the leading features were cancer-related genes. 
For example, the TMPRSS4 gene is an emerging potential therapeutic target in cancer ~\cite{de2015tmprss4}.
Moreover, tensin4 expression showed prognostic relevance in gastric cancer \cite{sakashita2008prognostic}.
Furthermore, E2F1-initiated transcription of PRSS22 promoted breast cancer metastasis by cleaving ANXA1 and activating the FPR2/ERK signaling pathway \cite{song2022e2f1}.
In addition, down-regulation of FXYD3 expression was observed in lung cancers \cite{okudela2009down}.
Moreover, ST14 gene expression affected breast cancer \cite{kauppinen2010st14, wang2009st14, dai2021gene}.
In addition, RAB25 has been implicated in various cancers, with reports presenting it as both an oncogene and a tumor-suppressor gene \cite{mitra2012rab25, wang2017rab25}. Long intergenic non-coding RNA 00324 promoted gastric cancer cell proliferation by binding with HuR and stabilizing FAM83B expression \cite{zou2018long}. CBLC expression was found to be higher in breast cancer tissues and cells than in normal
tissues and cells \cite{li2022cblc}. Furthermore, Serpin B5 was shown to be a CEA‐interacting biomarker for colorectal cancer \cite{baek2014serpin}. PKP3 interactions with the MAPK-JNK-ERK1/2-mTOR pathway regulated autophagy and invasion in ovarian cancer \cite{lim2019pkp3}. In addition, LAD1 expression was associated with the metastatic potential of colorectal cancer cells \cite{moon2020lad1}. ELF3 was found to be a negative regulator of epithelial--mesenchymal transition in ovarian cancer cells \cite{yeung2017elf3}. Moreover, the expression patterns of CEACAM5 and CEACAM6 were observed in primary and metastatic cancers \cite{blumenthal2007expression}.
Grhl2 determined the epithelial phenotype of breast cancers and promoted tumor progression \cite{xiang2012grhl2}.
KLF5 promoted breast cancer proliferation, migration, and invasion, in part by up-regulating the transcription of TNFAIP2 \cite{jia2016klf5}. Finally, genetic predisposition to colon and rectal adenocarcinoma was found to be mediated by a super-enhancer polymorphism coactivating CD9 and PLEKHG6~\cite{ke2020genetic}.

\subsubsection{Visualization Analysis}
We further performed a visual analysis to measure the performance of Gene-MOE. We randomly selected 1,000 patients from the test set to perform this evaluation. The real gene expression of test patients was first fed into Gene-MOE to generate the reconstruction expression. Then, we used TSNE and UMAP to perform the visualization and evaluated the similarity of these two distributions. Figure \ref{fig:feature}(c) shows the visualization result using these two methods. According to Figure \ref{fig:feature}(c), the real gene distribution perfectly coincided with the reconstructed gene distribution, which proves that the Gene-MOE model can perform reconstruction of the input genes more accurately based on the input real genes. Furthermore, Gene-MOE completed the reconstruction based on the low-dimensional feature obtained by the backbone model, which further reinforces that Gene-MOE can learn rich feature representation by the backbone network.

\begin{table*}[t]
\caption{Comparison of Concordance Index on 14 cancer types using four distinct models}
\begin{tabular*}{\textwidth}{@{\extracolsep\fill}lcccccccccccccc@{\extracolsep\fill}}
\toprule
& \multicolumn{1}{@{}c@{}}{BLCA}& \multicolumn{1}{@{}c@{}}{BRCA} & \multicolumn{1}{@{}c@{}}{HNSC} & \multicolumn{1}{@{}c@{}}{KIRC} & \multicolumn{1}{@{}c@{}}{LGG} & \multicolumn{1}{@{}c@{}}{LIHC} & \multicolumn{1}{@{}c@{}}{LUAD} & \multicolumn{1}{@{}c@{}}{LUSC} & \multicolumn{1}{@{}c@{}}{OV} & \multicolumn{1}{@{}c@{}}{SARC} & \multicolumn{1}{@{}c@{}}{CESC} & \multicolumn{1}{@{}c@{}}{STAD} & \multicolumn{1}{@{}c@{}}{COAD} & \multicolumn{1}{@{}c@{}}{UCEC}  \\
\midrule
Baseline &   0.643   &  0.692    &   0.632   & 0.728     &  0.857   &  0.691    &  0.649    &  0.558 & 0.590    & 0.722   &  0.720    &  0.607    &  0.674    &    0.699        \\
MOE & 0.670     &  0.711    &  \textbf{0.664}    & 0.729     & 0.848    & 0.700     & 0.663     &  0.596    & 0.616   & 0.733     &  0.720    & 0.641     & 0.705     & 0.712     \\
MOAE & 0.664     & 0.700     &  0.656    & \textbf{0.730}     & \textbf{0.850}    & \textbf{0.713}     & 0.660     &  \textbf{0.599}    &  0.613  &  0.736    & 0.733     & \textbf{0.728}     & 0.684     &  \textbf{0.719}    \\
pre-train    & \textbf{0.674}     & \textbf{0.718}     &  0.649    & \textbf{0.730}     & \textbf{0.850}    &  0.708    &  \textbf{0.666}    & 0.590     & \textbf{0.616}   & \textbf{0.737}     & \textbf{0.738}     & 0.630   & \textbf{0.715}     & 0.716    \\
\bottomrule
\end{tabular*}
\label{tab:analysis}
\end{table*}

\begin{table}[t]
\caption{Comparison of classification metrics using four distinct models}%
\begin{tabular*}{0.95\columnwidth}{@{\extracolsep\fill}lcccc@{\extracolsep\fill}}
\toprule
  & Accuracy & Precision & Recall & F1-Score \\ \midrule
Baseline         & 0.9304 &   0.9235 & 0.8721       &  0.8799       \\
MOE         & 0.9417 &   0.9214 & 0.9063       &  0.9098        \\
MOAE       &  \textbf{0.9584}        &  0.9469     &  0.9349      &   \textbf{0.9392}       \\
pre-train & 0.9580    &  \textbf{0.9554}     & \textbf{0.9374}  & 0.9333         \\ \bottomrule
\end{tabular*}
\label{tab:class}
\end{table}

\subsection{Ablation Studies}
In this section, we extend our analysis by presenting ablation experiments conducted to evaluate the performance of the model proposed in this paper. Four distinct models were constructed for this purpose: 1) the baseline model comprising two FCN layers, 2) the model incorporating MOE by replacing the FCN layers, 3) the Gene-MOE model, and 4) the pre-trained Gene-MOE model. Subsequently, survival analysis and cancer classification tasks were performed using these four models. The results of these tasks are presented in Tables \ref{tab:analysis} and \ref{tab:class}.

\begin{figure}[!t]
     \centering
     \includegraphics[width=\columnwidth]{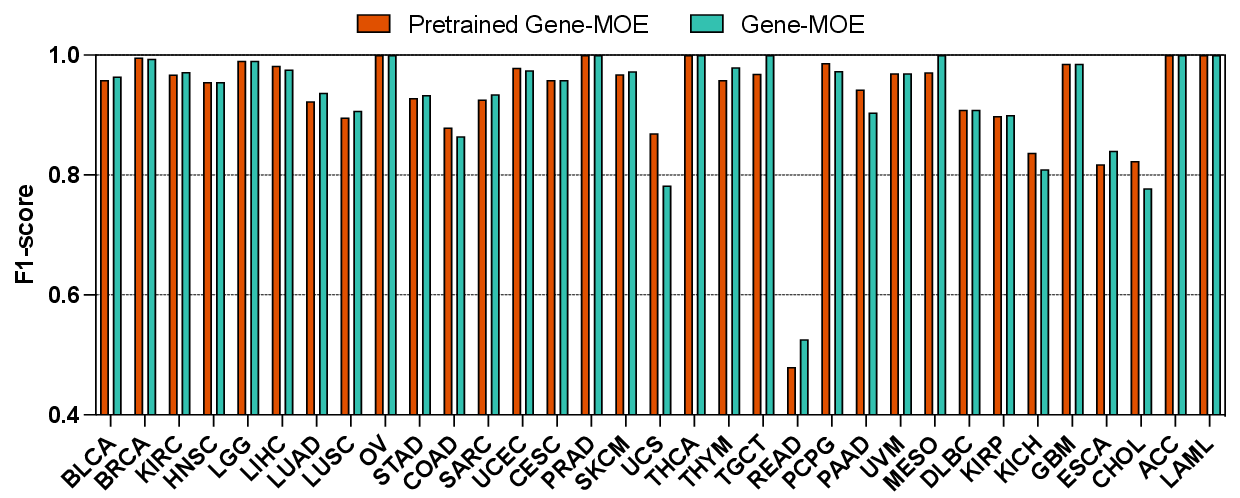}
     \caption{Comparison of F1-score on 33 cancer types using pre-trained Gene-MOE and Gene-MOE.}
     \label{fig:f1}
\end{figure}

\subsubsection{Performance of MOE}

By comparing survival analysis result in Table \ref{tab:analysis}, we found the model with the MOE layer could achieve better Concordance results than baseline on 12 cancer types. Moreover, in Table \ref{tab:class}, the model with MOE showed better accuracy, and F1-score.
These results indicate that using the MOE layer can increase the performance of genomic analysis, which proves our theory that using MOE can make a model learn rich features during the feature extraction process.

\subsubsection{Performance of MOAE}

By comparing the Concordance Index for model with MOE and the Gene-MOE model in Table \ref{tab:analysis}, we observed that Gene-MOE outperformed the model with MOE on nine cancer types. 
Furthermore, in Table \ref{tab:analysis}, we found that Gene-MOE performed better in accuracy, recall, precision, and F1-score.
These findings prove that the MOAE model can further improve the accuracy of the model by improving the ability to learn deep semantic correlated features.


\subsubsection{Performance of Pre-training}

By comparing the performance of the Gene-MOE and pre-trained Gene-MOE models in Table \ref{tab:analysis}, we found that the pre-trained Gene-MOE model performed better in eight cancer types. At the same time, the Gene-MOE model demonstrated the same performance on KIRC and LGG datasets. These results reveal the effectiveness of pre-training, particularly in enhancing the performance of most survival analysis models.
Further comparison of Gene-MOE and pre-trained Gene-MOE in Table \ref{tab:class} showed that that the pre-trained Gene-MOE model performed better in the precision and recall, but the accuracy and F1-score were lower than that of Gene-MOE. By further compare the F1-score of these two model in Figure \ref{fig:f1}, we found that the pre-trained model performed less effective on cancer types with few samples, such as the READ dataset. This discrepancy can be attributed to the fact that the pre-training stage primarily captures common cancer features, potentially missing certain characteristics for cancer types with small sample sizes. Consequently, during the fine-tuning stage, the model tends to favor learning patterns from cancer types with larger sample sizes.

\section{Conclusion} \label{conclusion}

In this work, we developed a sparsely gated model called Gene-MOE, which extensively leverages MOE layers to further deepen
the ability to extract the deep correlation features of high-dimensional genes. Furthermore, we proposed a novel MOAE module to explore the deep semantic associations between high-dimensional genetic features.
Finally, we designed novel pre-training strategies including data
augmentation, self-supervised learning, and new loss functions to further improve the performance of Gene-MOE.
The results show that Gene-MOE could achieve the best performance on cancer classification and survival analysis, indicating its strong potential for use in those applications.
Currently, however, Gene-MOE has some limitations. During the pre-training stage, Gene-MOE focuses on cancer types with larger sample sizes, resulting in insufficient fitting of small sample datasets. Furthermore, the amount of existing data is insufficient, which leads to over-fitting issues in survival analysis and cancer classification tasks.
In our future work, we aim to gather more genetic data for model training and to optimize the model training performance.



\section{Acknowledgment}

This work was supported by the National Key Research and Development Project of China under Grant 2021YFA1000103 and 2021YFA1000100, the National Natural Science Foundation of China under Grant 62372469, 61972416, 62272479 and 62202498, the Taishan Scholarship under Grant tsqn201812029, the Foundation of Science and Technology Development of Jinan under Grant 201907116, the Shandong Provincial Natural Science Foundation under Grant ZR2022LZH009 and ZR2021QF023, the Fundamental Research Funds for the Central Universities under Grant 21CX06018A, the Spanish project under Grant PID2019-106960GB-I00, the Juan de la Cierva under Grant IJC2018-038539-I.

\bibliographystyle{IEEEtran}
\bibliography{IEEEabrv, reference}



\end{document}